\documentclass[11pt,a4paper]{article}
\usepackage[hyperref]{acl2020}
\usepackage{times}
\usepackage{latexsym}

\usepackage{amsmath,bm}
\usepackage{mathrsfs}
\usepackage{multirow}
\usepackage{CJKutf8}
\usepackage{verbatim}
\usepackage{amssymb}
\usepackage{algorithm}
\usepackage{algorithmic}
\usepackage{graphicx}
\usepackage{color}
\usepackage{subfigure}
\usepackage{hyperref}
\usepackage{enumitem}
\usepackage{makecell}
\usepackage[bottom]{footmisc}
\usepackage{textcomp}
\usepackage{graphics} 
\usepackage{adjustbox}

\hypersetup{colorlinks=true,linkcolor=blue,urlcolor=blue}
\newcommand{\langl}{\begin{picture}(4.5,7)
\put(0,2.5){\rotatebox{45}{\line(1,0){5.5}}}
\put(0,2.5){\rotatebox{315}{\line(1,0){5.5}}}
\end{picture}}
\newcommand{\rangl}{\begin{picture}(4.5,7)
\put(-2.5,2.5){\rotatebox{135}{\line(1,0){5.5}}}
\put(-2.5,2.5){\rotatebox{225}{\line(1,0){5.5}}}
\end{picture}}

\definecolor{deepgray}{HTML}{708090}

\newcommand{\angl}[1]{\langl \nolinebreak \textit{#1} \nolinebreak \rangl}
\newcommand{\eos}{\angl{eos}}
\newcommand{\pad}{\angl{pad}}
\newcommand{\unk}{\angl{unk}}
\newcolumntype{C}[1]{>{\centering}p{#1}}
\makeatletter
\newcommand\footnoteref[1]{\protected@xdef\@thefnmark{\ref{#1}}\@footnotemark}
\makeatother

\usepackage{microtype}

\aclfinalcopy % Uncomment this line for the final submission
\title{CoTK: An Open-Source Toolkit for \\
Fast Development and Fair Evaluation of Text Generation}

\author{
Fei Huang, Dazhen Wan, Zhihong Shao, Pei Ke\\
\textbf{Jian Guan, Yilin Niu, Xiaoyan Zhu, Minlie Huang} \\
Institute for Artificial Intelligence, 
State Key Lab of Intelligent Technology and Systems \\
Beijing National Research Center for Information Science and Technology \\
Department of Computer Science and Technology, Tsinghua University, Beijing 100084, China \\
  {\tt \{f-huang18, wadz19, szh19, kp17, j-guan19,} \\
  {\tt nyl18\}@mails.tsinghua.edu.cn}\\
  {\tt \{zxy-dcs, aihuang\}@tsinghua.edu.cn} \\
}

\date{}

\begin{document}

\maketitle
\begin{abstract}
In text generation evaluation, many practical issues, such as inconsistent experimental settings and metric implementations, are often ignored but lead to unfair evaluation and untenable conclusions.
%Automatic evaluation of text generation is not a trivial problem, where many practical issues can lead to unfair comparisons between models.
%
We present CoTK, an open-source toolkit aiming to support fast development and fair evaluation of text generation.
%Instead of focusing on the model implementation as most existing toolkits do
In model development, CoTK helps handle the cumbersome issues, such as data processing, metric implementation, and reproduction. It standardizes the development steps and reduces human errors which may lead to inconsistent experimental settings.
In model evaluation, CoTK provides implementation for many commonly used metrics and benchmark models across different experimental settings. As a unique feature, CoTK can signify when and which metric cannot be fairly compared.
%As a unique advantage, CoTK provides hash code to identify datasets and metrics under different settings, which avoid unfair comparisons between models.
%
%Moreover, CoTK helps share the models with the community, where other researchers can reproduce the results quickly.
%We also provide resources and benchmark models to enhance usability. 
%For an easy start, we provide benchmarks and baselines for some tasks.
%
We demonstrate that it is convenient to use CoTK for model development and evaluation, particularly across different experimental settings.
\end{abstract}

\setlength{\textfloatsep}{10.0pt}
\setlength{\intextsep}{9.0pt}
\setlength{\abovedisplayskip}{6pt}
\setlength{\belowdisplayskip}{6pt}
\setlength{\skip\footins}{6pt} %11pt

\section{Introduction}

Neural text generation, as a key but challenging task in NLP, has been widely studied recently. 
Text generation has been applied to various scenarios, such as dialog generation \cite{vinyals2015conversational}, story generation \cite{story2016roemmele}, machine translation \cite{seq2seq2014sutskever}, text summarization \cite{summarization2015rush} and image captioning \cite{imagecaption2015vinyals}.

Novel methods for text generation are constantly developed, but the evaluation of text generation is much less touched~\cite{huang2019challenges}. Even worse, it is common that the results provided by prior models contradict one another, and thus it is hard to identify state-of-the-art models for some task. After thorough investigation of existing open-source projects on text generation, we observe two common problems in model evaluation.

\begin{figure}[!t]
  \centering
  \includegraphics[width=\linewidth]{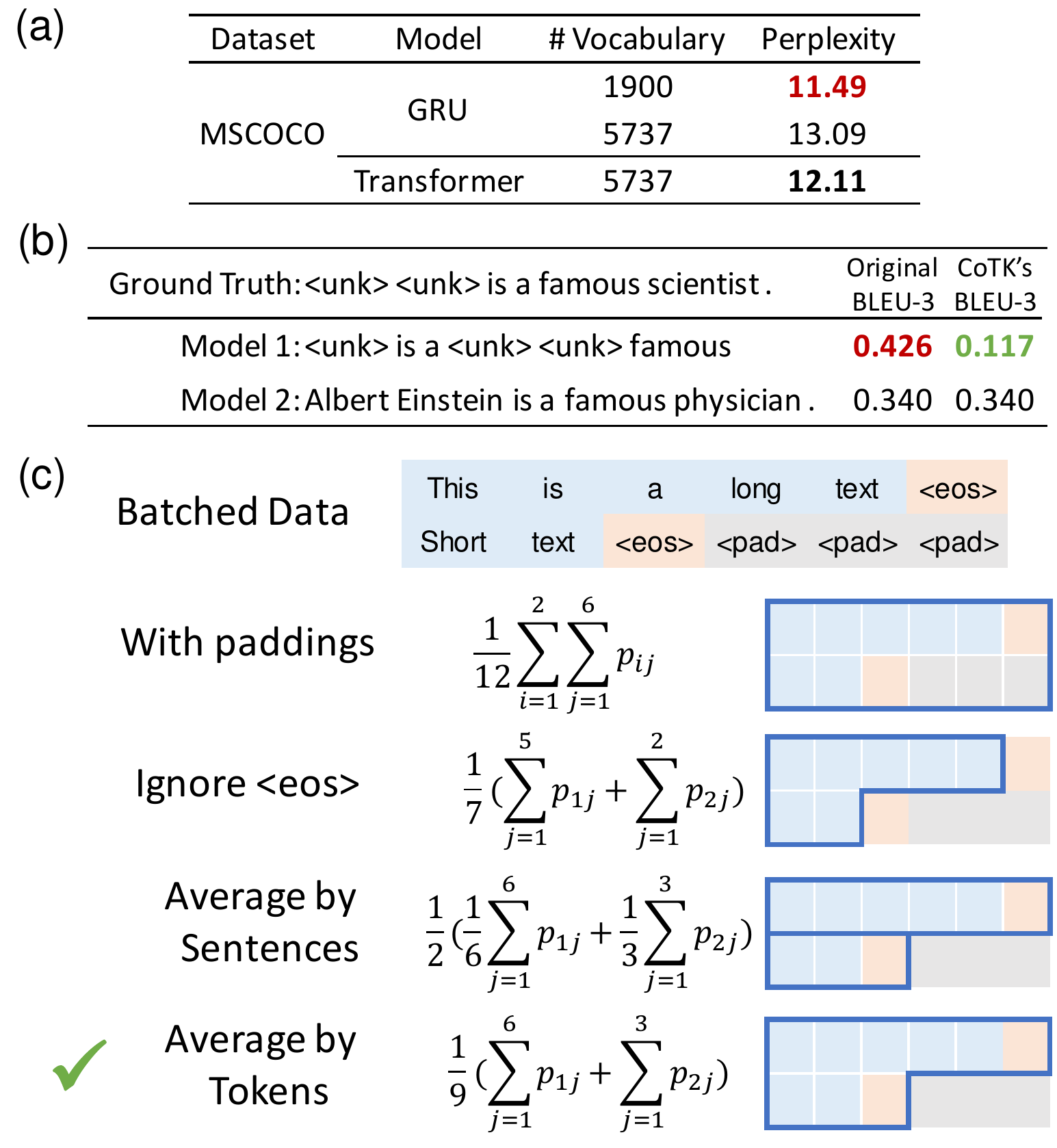}
  \caption{Examples of problems leading to inconsistent results. (a) Transformer is better than GRU when compared in the same setting. If GRU is tested on a smaller vocabulary, the unfair comparison may lead to a wrong conclusion. (b) If the ground truth contains \unk (unknown tokens), the original BLEU-3 may prefer the worse sentence (the first one). (c) The different implementations of perplexity lead to inconsistent results, where \eos means tokens at the end of sentences and \pad means paddings. We present the formulations of log perplexity, where the negative log probability of the $j$-th token in the $i$-th sentence $p_{ij} = -\log P(x_{ij}|x_{i, <j})$.}
  \label{fig:intro-example}
  %\vspace{-0.3em}
\end{figure} 

One problem is that models are tested on different datasets and with different experimental settings, thereby making these models incomparable. 
%There are few widely accepted benchmarks in this area. 
Even on the same dataset, the settings used in evaluation, such as the data split (training/test), the method of tokenization and the size of vocabularies, differ unconsciously, thereby leading to unfair comparison between models and untenable conclusions. As shown in Figure \ref{fig:intro-example} (a), the subtle difference in vocabularies can completely change the results. However, uncovering the differences among these settings can be extremely difficult, which makes results hardly reproducible.

The other problem lies in that the metrics of text generation are rather complicated, leading to inconsistency among different implementations. For example, as shown in Figure \ref{fig:intro-example} (b), the truncation of the vocabulary brings \unk (unknown tokens) into ground truth, and the original BLEU metric \cite{bleu2002papineni} favors the sentence containing more \unk. %The same problem can also happen on other metrics. 
In (c), we present several implementations of perplexity \cite{perplexity1992brown}, leading to very different results.

These problems severely prevent us from comparing different models fairly, reproducing existing models and implementing new models. There is a heavy burden in checking the details of experimental settings, metric implementations, and more. To this end, we develop Conversational Toolkit (CoTK), an open-source~\footnote{CoTK is available at \href{https://github.com/thu-coai/cotk}{https://github.com/thu-coai/cotk} with Apache License 2.0.} toolkit as a python package. CoTK is mainly developed for open-domain conversation generation, but other tasks of text generation are also supported. CoTK is designed to achieve two goals:
\begin{itemize}[leftmargin=1em]
    \setlength{\itemsep}{0ex}
    \setlength{\parskip}{2px}
    \vspace{-0.1em}
    \item Empowering fast development. CoTK helps handle the cumbersome issues in data loading, processing, evaluation, and reproduction, so that the researchers can concentrate on the most creative part, i.e., the implementation of novel models.
    
    \item Empowering fair evaluation. CoTK is specially designed for ensuring fair comparison, where a unique hash code can signify whether the experimental results are comparable.
    \vspace{-0.1em}
\end{itemize}

\noindent In CoTK, we provide:
\begin{itemize}[leftmargin=1em]
    \setlength{\itemsep}{0ex}
    \setlength{\parskip}{2px}
    %\vspace{-0.6em}
    \vspace{-0.1em}
    \item Data loaders. We build data loaders for common text generation tasks, where the data loaders handle the whole procedure before sending data into the models, including reading files, processing, and packing sample batches.
    
    \item Metrics. CoTK covers commonly used metrics in text generation tasks. Each evaluation result will be tagged with a hash code, which can be used to verify the fairness of comparisons.
    
    \item A tool for publication and reproduction. CoTK can track the code and experimental environment, which enables researchers to publish models or reproduce others' results conveniently.
    
    \vspace{0.5em}
    
    \item Resources and benchmark models. We collect some public datasets and commonly used benchmark models, which facilitate development of new models and comparison with existing ones.
\end{itemize}

\section{Design and Structure}

\begin{figure}[!t]
  \centering
  \includegraphics[width=0.95\linewidth]{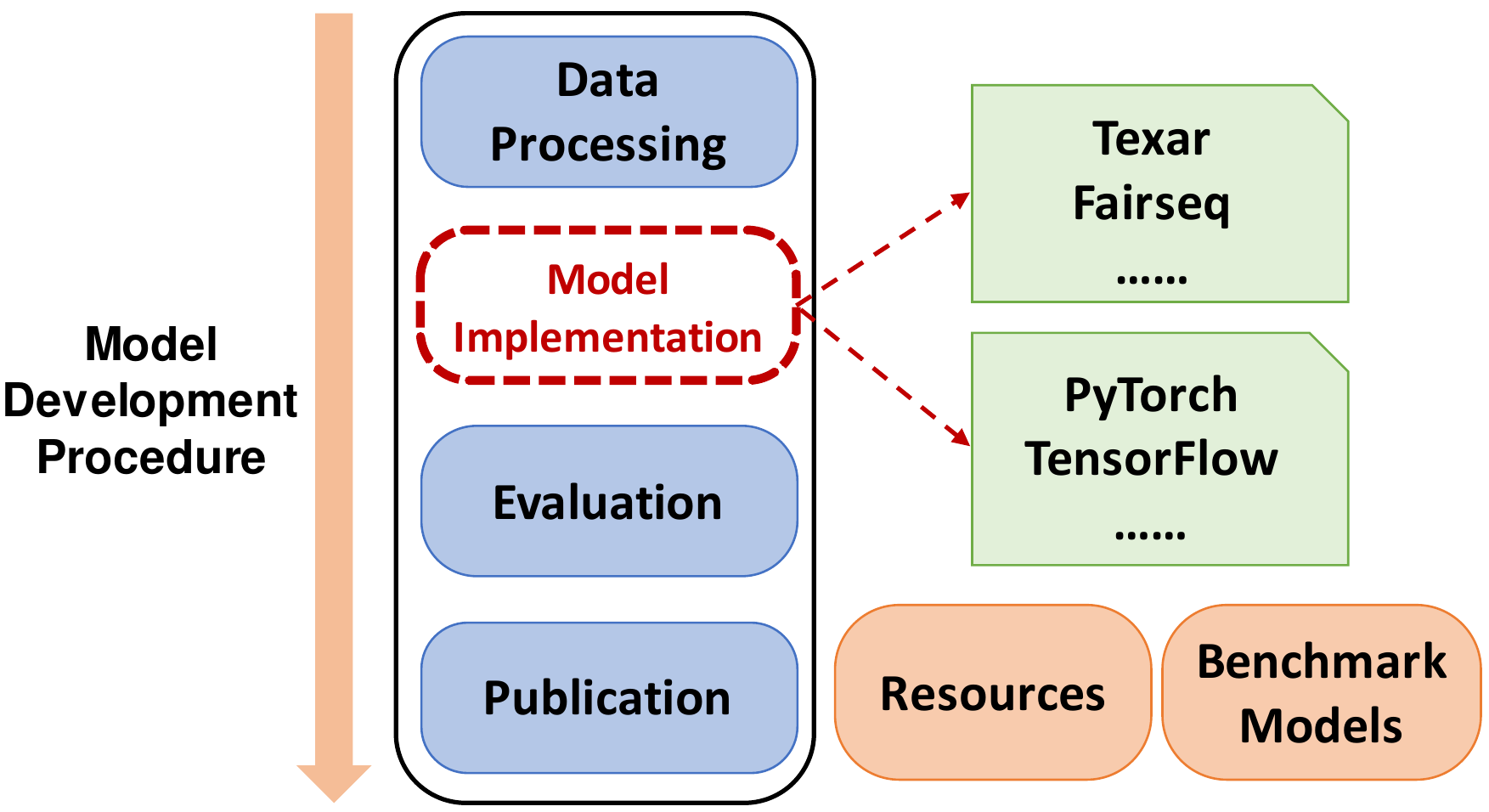}
  \caption{CoTK's design concepts. We provide tools for data processing, evaluation and model publication, as well as resources and benchmark models. CoTK is compatible with many existing toolkits that provide various model implementations.} %CoTK can is compatible with many other toolkits that can complement our toolkit in the model implementation so that we can provide full support throughout the model development procedure.
  \label{fig:overiew}
  \vspace{-0.5em}
\end{figure}

CoTK aims at supporting researchers through the entire lifetime of model development. As shown in Figure \ref{fig:overiew}, we divide the development procedure into four steps: data processing, model implementation, evaluation, and publication. 
%As aforementioned, 
%To make researchers concentrate on the model implementation, 
CoTK characterizes itself in three aspects: data loaders for data processing, metrics for evaluation, and a tool for publication and reproduction. For model implementation, CoTK is compatible with many existing toolkits. That is, researchers can be supported in data processing, evaluation, and publication from CoTK while implementing models with other toolkits, such as Texar \cite{texar2019} and Fairseq \cite{ott2019fairseq}, in PyTorch \cite{pytorch2019}, TensorFlow \cite{tensorflow2016} or other deep learning frameworks.
%%%其它工具集的模型，如何才能用到cotk上，需要做什么？直接拿着COTK的接口测试就可以
%, which complement our toolkit in the model implementation.%, as described in Section \ref{sec:related}.
%We also provide processed datasets and benchmark models, where researchers can make a quick start with CoTK.

\subsection{Data Loader}

\label{sec:dataloader}

Data loader helps users prepare data for deep learning models. Following user-specified settings, data loader can read files, make tokenization, build vocabularies and pack sentences to mini-batches.

\vspace{0.5em}

\noindent \textbf{Tasks} \\
%
%A task specifies the format of the data file. 
A task is specified by the configurations shown in Table \ref{tab:task}.
We mainly support text generation (without input) \cite{text2011sutskever}, single-turn dialog generation \cite{vinyals2015conversational} and multi-turn dialog generation \cite{hred2015sordoni}. By assembling different types of input and output, CoTK can be easily extended to various tasks, such as machine translation \cite{seq2seq2014sutskever}, 
%text summarization \cite{summarization2015rush}, 
controllable conversation generation \cite{ecm2018zhou}. 
%We present some examples of supported tasks in Table \ref{tab:task}.

%A task specifies the format of the data file, where we mainly support text generation (without input) \cite{text2011sutskever}, single-turn dialog generation \cite{vinyals2015conversational}, and multi-turn dialog generation \cite{hred2015sordoni}. 
%Since several types of data (named as data fields) may be shared among different tasks, we further decompose the data loaders into modularized parts. 
%To extend our support for new tasks, which can be composed of several data fields. so our data loaders are able to support other text generation tasks, such as machine translation \cite{seq2seq2014sutskever}, text summarization \cite{summarization2015rush}, controllable conversation generation \cite{ecm2018zhou}. We present the data fields of different tasks in Table \ref{tab:task}.

\begin{table} [!t]
\centering
\small
\setlength{\tabcolsep}{0.8mm}{
\begin{tabular}{llll}
\hline
Task & Configuration \\
\hline
Text Generation (w/o input) & $\varnothing \rightarrow$ Sentence \\
Single-Turn Dialog & Sentence $\rightarrow$ Sentence \\
Multi-Turn Dialog & Context $\rightarrow$ Sentence \\
Machine Translation & Sentence $\rightarrow$ Sentence \\
%Text Summarization & Sentence $\rightarrow$ Sentence \\
Controllable Generation & (Sentence, Label) $\rightarrow$ Sentence \\
\hline
\end{tabular}
}
\caption{Examples of tasks supported by CoTK. Configuration is described by Input $\rightarrow$ Output. $\varnothing$ denotes no input, and Context means a sequence of utterances.}
\label{tab:task}
\vspace{-0.5em}
\end{table}

\vspace{0.5em}

\noindent \textbf{Tokenization} \\
The means of tokenization are usually ignored but can largely affect the experimental results. %, as shown by the example in Section \ref{sec:diff-tokenize}.
We provide widely used Puckt tokenizer~\cite{Punkt} as well as tokenizers for GPT-2~\cite{gpt22019radford} and other pretraining models.
%We provide the widely used Punkt tokenizer \cite{punkt} as our default tokenizer. However, aware of the rising of pre-trained models, we further provide pre-trained tokenizers for BERT\cite{bert2019}, GPT-2\cite{gpt22019radford}, and more. The implementations of tokenizers are from NLTK \cite{loper2002nltk} and HuggingFace's Transformers \cite{Wolf2019HuggingFacesTS}.

\vspace{0.5em}

\noindent \textbf{Vocabulary} \\
%
%In order to reduce the parameters of the generation model, it is a common method to filter out rare words in the vocabulary. However, we should be careful when comparing the models trained on different sets of vocabulary. A model trained on a dataset with a smaller vocabulary usually achieves lower perplexity, but it does not indicate the superiority of the model.
It is a common choice to filter out rare words in the vocabulary. However, fair comparison among the models trained with different vocabularies is not trivial, as shown by the example in Figure \ref{fig:intro-example} (a). %To tackle the issue, we formally define our vocabularies.
To this end, we split a vocabulary into two parts:
%We split a vocabulary into three parts, valid words ($F$), invalid words ($R$), and unknown words ($U$) as follows:

%Figure \ref{fig:words} describes the split and the functions. 

%$F$ are the frequent words in the training data, and $F \cup R$ covers the test data. This definition is similar to \citet{copynet2017he}, but their split changes for each sample while ours is the same for the whole dataset.  
%Valid words are the frequent words from the **training** set. 

\begin{comment}
\begin{figure}[!b]
  \centering
  \includegraphics[width=\linewidth]{words.pdf}
  \caption{The split of the vocabularies and their functions. $F$ is valid words and $R$ is invalid words. Copyable means invalid words can be generated via copy mechanism \cite{copynet2017he}. }
  \label{fig:words}
  \vspace{-0.3em}
\end{figure} 
\end{comment}

%\begin{comment}
\begin{itemize}[leftmargin=1em]
    \setlength{\itemsep}{0ex}
    \setlength{\parskip}{2px}
    \vspace{-0.6em}
    \item Frequent vocabulary ($F$). Frequent vocabulary contains frequent words from the training set. It is the vocabulary used by most models.% They can be generated by models.% and evaluated by metrics.
    %from the \textbf{training} set. Models can generate the valid words, and the valid words will be evaluated by metrics.
    \item Rare vocabulary ($R$). Rare vocabulary contains the remaining words from the training and the test set, which cannot be generated by most models except the copy mechanism \cite{copynet2017he}. Note that $R$ and $F$ have no intersection. % and can be evaluated by metrics.
    %\item Invalid words. Invalid word comes from the \textbf{training}, \textbf{development}, or \textbf{test} set but it must not be a valid word. Only some models like CopyNet \cite{} can generate the invalid words via copying. The invalid words will also be evaluated by metrics.
   % \item Unknown words ($U$). Unknown words are all the other words not appeared in the dataset. They cannot be generated by models and will be ignored when evaluated by metrics. This avoids some degenerated conditions as showed in Figure \ref{fig:intro-example} (b).
    %\item Unknown words. Unknown words are all the other words which are not valid or invalid words. Even a word not appeared in the \textbf{training}, \textbf{development}, or \textbf{test} set can be a unknown word. The model cannot generate unknown words, and the unknown words will also not be evaluated by metrics.
    \vspace{-0.5em}
\end{itemize}
%\end{comment}

In the training stage, models only see $F$ and they regard all words not in $F$ as \unk. In the test stage, models are evaluated on $F \cup R$. Although rare words cannot be generated by most models, they are crucial for evaluation.
%Based on the definition, 
Our metrics are designed to achieve fair comparison as long as $F \cup R$ does not change. %This enable us to compare models trained on various $F$.
Supposing that two models are trained with different frequent vocabularies, $F_A$ and $F_B$ for instance, they can be fairly compared by adjusting the rare vocabularies to keep $ F_A \cup R_A = F_B \cup R_B$.

%Unknown words $U$ can be an infinite set, but invalid words $R$ should be a finite set, because $|R|$ will be used by some metrics. The definition of $U$ is useful when evaluating the model without specifying the whole test set before evaluation, for example, in an online learning algorithm. Please refer to Section \ref{sec:metric} for how our metrics process different types of words.

\vspace{0.5em}

\noindent \textbf{Hash Code for Data Loader} \\
Since it is difficult to track the differences among various data loaders, CoTK provides hash codes to identify each part of the data loader including the input data, vocabularies and settings (shown in Table \ref{tab:data-hash}).
%As tracking the differences in data loader can be difficult, CoTK provides hash codes to identify data loaders with different data, vocabularies, or settings as shown in Table \ref{tab:data-hash}.
For example, if two data loaders have the same \textit{General Hash} code, their data, vocabularies and settings are guaranteed to be the same. This is implemented by computing SHA-256 given the corresponding parts of data loaders as input. A usage case is presented in Section \ref{sec:hashcode-example}.

% This feature helps researchers to find whether the differences come from the data or the models.
%As aforementioned, the settings of datasets are crucial to the result, but tracking all the details is difficult. To tackle the problems, CoTK provides hash codes to identify data loaders in different settings. If two data loaders have the same hash code, their settings and data are guaranteed to be the same. This feature helps researchers to find whether the differences come from the data or the models.
%To further expand the functionality, we design hash codes for identifying different parts of settings, as shown in Table \ref{tab:data-hash}.

\begin{table} [!t]
\centering
\small
\setlength{\tabcolsep}{3.0mm}{
\begin{tabular}{ll}
\hline
Hash Code & Object to be Identified \\
\hline
Raw Data Hash & Raw Text Data \\
Data Hash & Tokenized Data \\
Vocab Hash & Vocabulary \\
Setting Hash & Settings \\
General Hash & All Above \\
\hline
\end{tabular}
}
\caption{Hash codes used to identify different parts of data loaders including data, vocabulary and settings.}
\label{tab:data-hash}
%\vspace{-1.5em}
\end{table}

\subsection{Metric}
\label{sec:metric}

CoTK covers commonly used metrics in text generation tasks, as shown in Table \ref{tab:metrics}.

\begin{table} [!htp]
\centering
\small
\setlength{\tabcolsep}{1.0mm}{
\begin{tabular}{l}
\hline
\textbf{Text Generation (Without Input)} \\
\hline
Perplexity \cite{perplexity1992brown}\\
Self-BLEU \cite{texygen2018} \\
Forward / Backward BLEU \cite{irl2018}\\
Forward / Reverse Perplexity \cite{arae2018} \\
\hline
\textbf{Dialog Generation} \\
\hline
%Perplexity \cite{perplexity1992brown}\\
BLEU \cite{bleu2002papineni}\\
Distinct N-gram \cite{distinct2016li} \\
BOW Embedding \cite{forgues2014bootstrapping} \\
%BLEU Precision / Recall \cite{cvae2017} \\
%BOW Embedding Precision / Recall \cite{cvae2017} \\
\hline
\textbf{Machine Translation \& Text Summarization} \\
\hline
%Perplexity \cite{perplexity1992brown}\\
%BLEU \cite{bleu2002papineni}\\
ROUGE \cite{rouge2014lin}\\
METEOR \cite{meteor2005banerjee} \\
\hline
\end{tabular}
}
\caption{Part of the metrics supported by CoTK.}
\label{tab:metrics}
\vspace{-0.5em}
\end{table}

In CoTK, metrics are implemented to achieve fair comparison among models in different experimental settings.
%modified for tackling the Out-Of-Vocabulary issue, as shown in Figure \ref{fig:intro-example} (b). 
We will take perplexity and BLEU as examples to introduce our implementation. %Please refer to Section \ref{sec:dataloader} for different types of vocabulary before we explain the examples.

\vspace{0.5em}

\noindent \textbf{Example: Perplexity} \\
The original perplexity is calculated as
\begin{align}
perplexity = \exp(-\frac{1}{N}\sum_{i=1}^{N} \log p_{\theta}(w_i)),  \notag
\end{align}
where $w_i$ is the $i$-th token in the ground truth, and $p_{\theta}$ is given by the model $\theta$. Supposing two models are trained with different frequent vocabularies, denoted as $F_A$ and $F_B$ respectively, where $F_A \subset F_B$, and there exists $w_i \notin F_A$ but $w_i \in F_B$. Then the model B should predict the exact $w_i$, but the model A only need to predict a \unk. It is unfair to compare the two models based on the original perplexity, as shown in Figure \ref{fig:intro-example} (a).

%This problem has been addressed by \citet{ahn2016upp}, where they introduce a new metric named \textit{unknown penalized perplexity}. 

Similar to \citet{ahn2016upp}, we distribute the probability of \unk evenly to the rare words:
\begin{gather}
perplexity = \exp(-\frac{1}{N}\sum_{i=1}^{N} t_i),  \notag \displaybreak[0]\\
t_i =  \left\{
    \begin{array}{lr}
        \log p_{\theta}(w_i) & \text{if } w_i \in F, \\
        \log (p_{\theta}(unk) / |R|) & \text{if } w_i \in R,\\
        %0 & \text{if } w_i \in U,
    \end{array} \right. \notag \displaybreak[0]
%T = \sum_{i=1}^{N} 1[w_i \in (F \cup R)], \notag
\end{gather}
where $F$, $R$ are frequent vocabulary and rare vocabulary respectively. %For most models, $p(w_i)=0$ if $w_i \in R$, except the models which can generate $w_i$ via copy mechanism \cite{copynet2017he}.
This method converts the predicted probability distribution over $F \cup \{unk\}$ to a distribution over $F \cup R$, so that the perplexity can always be fairly compared as long as $F \cup R$ keeps unchanged.

%Therefore, we are able to achieve a fair comparison between two models in different settings, as long as $V_A \cup I_A = V_B \cup I_B$. %no matter whether $V_A = V_B$ or copy mechanism is applied.

\vspace{0.5em}

\noindent \textbf{Example: BLEU} \\
The BLEU metric may be affected by two issues: different tokenizers bring different token sets; BLEU may favor sentences with \unk, as shown in Figure \ref{fig:intro-example} (b). 

In CoTK's BLEU, we first concatenate tokens for both hypotheses and references and then make tokenization again by Puckt tokenizer. This step standardizes the tokenization.
Then we count the matches of n-grams following the original BLEU, but we never match n-grams containing \unk. It is because \unk is not a real token and should be always regarded as mismatched.

\begin{figure}[!htp]
  \centering
  \includegraphics[width=\linewidth]{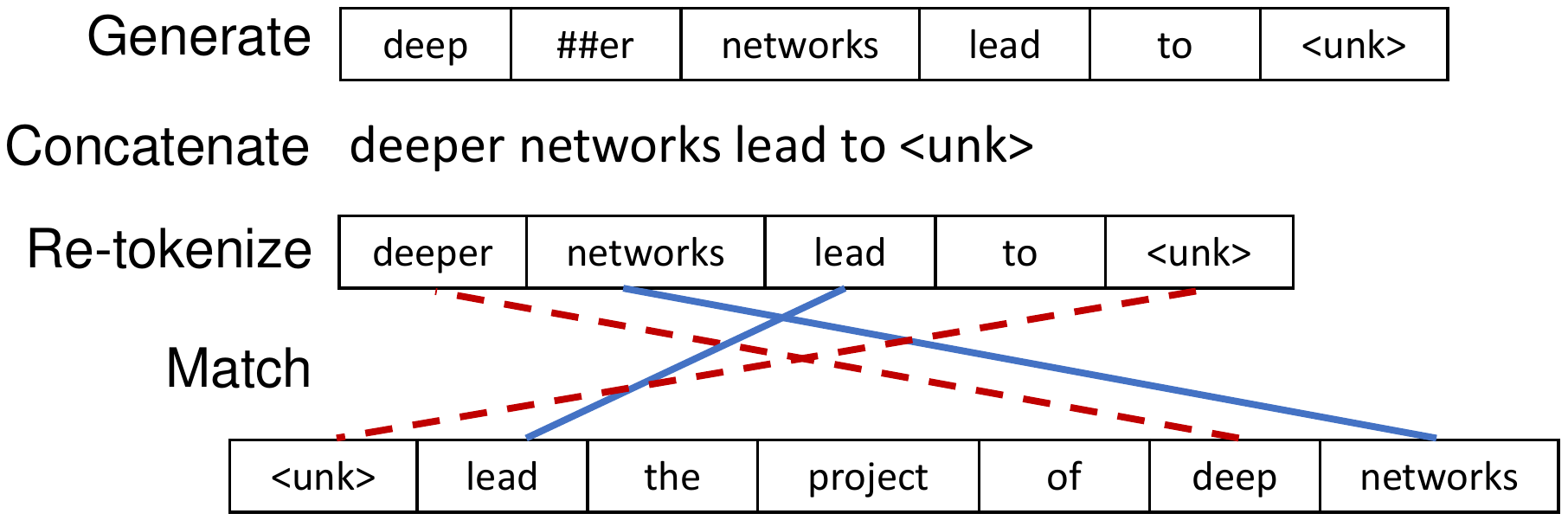}
  \caption{The steps of calculating BLEU in CoTK. Blue solid lines align matched pairs, and red dotted lines denote unmatched pairs.}
  \label{fig:bleu}
  \vspace{-0.3em}
\end{figure} 

This modification greatly extends applicability, which ensures fair comparison regardless of tokenization methods or vocabulary sets adopted by generation models.

\vspace{0.5em}

\noindent \textbf{Hash Code for Metric} \\
Hash codes generated for metrics can track the settings and the reference data, where two metric scores are comparable if and only if they have the same hash code. The implementation of the hash code in each metric can be different. For example, the hash code of perplexity computes the SHA-256 hash given the reference sentences, the frequent vocabulary, and rare vocabulary as input. However, the computation of the hash code for BLEU only uses the tokenized reference sentences as input, because BLEU does not rely on the vocabulary set for fair comparison.%We will present more showcases in Section \ref{sec:showcase}.
%because the range of fair comparison is different in these two metrics.

The hash codes has several advantages: It avoids human errors such as inconsistent settings; It saves researchers from memorizing the requirements of each metric for fair comparison. A case of usage is presented in Section \ref{sec:hashcode-example}.
%It is a convenient indicator to determine which values can be compared;

\begin{table} [!t]
\centering
\small
\setlength{\tabcolsep}{1.0mm}{
\begin{tabular}{cl}
\hline
Tasks                              & Resources     \\
\hline
\multirowcell{2}{Text Generation \\ (Without Input)}            & MSCOCO \cite{chen2015mscoco}        \\
& EMNLP2017 WMT\footnotemark \\
\hline
Single-Turn Dialog                 & OpenSubtitles \cite{opensubtitles2016tiedemann} \\
\hline
\multirow{2}{*}{Multi-Turn Dialog} & Ubuntu \cite{ubuntu2015lowe}        \\
 &SwitchBoard\footnotemark \cite{cvae2017} \\
\hline
\end{tabular}
}
\caption{Some datasets supported by CoTK.}
\label{tab:resources}
\vspace{-0.5em}
\end{table}

\addtocounter{footnote}{-2}
\addtocounter{footnote}{1}\footnotetext{Only monolingual corpus used. \url{http://statmt.org/wmt17/translation-task.html}}

\begin{table} [!t]
\centering
\small
\setlength{\tabcolsep}{1.0mm}{
\begin{tabular}{cl}
\hline
Tasks                               & Benchmark Models            \\
\hline
\multirowcell{4}{Text Generation \\ (Without Input)}    & GRU\cite{graves2013rnn,gru2014chung}                  \\
& Transformers\cite{vaswano17transformer}         \\
& GPT2-finetune\cite{gpt22019radford}\\
& VAE\cite{vae2014}        \\         
%& SeqGAN\cite{yu2017seqgan}               \\
\hline
\multirow{3}{*}{Single-Turn Dialog} & Seq2Seq-GRU\cite{seq2seq2014sutskever}          \\
                                    & Seq2Seq-Trans\cite{vaswano17transformer} \\
                                    & GPT2-finetune\cite{wolf2019transfertransfo} \\
\hline
\multirow{2}{*}{Multi-Turn Dialog}  & HRED\cite{hred2015sordoni}     \\
                                    & CVAE\cite{cvae2017}             \\   
\hline
\end{tabular}
}
\caption{Some benchmark models provided by CoTK.}
\label{tab:baselines}
\vspace{-0.5em}
\end{table}

\subsection{Publication and Reproduction}

\addtocounter{footnote}{1}\footnotetext{\href{https://catalog.ldc.upenn.edu/LDC97S62}{https://catalog.ldc.upenn.edu/LDC97S62}}

\label{sec:pub-repro}

To further improve reproducibility, we develop a tool that helps researchers publish their code and experimental results. 

\noindent \textbf{Publication}: If a user wants to share the results with the community, the user should follow a few steps: (1) Use the version control system git\footnote{\href{https://git-scm.com/}{https://git-scm.com/}} to track code updates. (2) Write code that generates a result file when executed. (3) Execute the code from our tool. These three steps track the code, the results, and the running environment. Then all these data can be uploaded to our website\footnote{\label{footnote:dashboard}\href{http://coai.cs.tsinghua.edu.cn/dashboard/}{http://coai.cs.tsinghua.edu.cn/dashboard/}} or GitHub\footnote{\href{https://github.com/}{https://github.com/}}, which are accessible to the community. We highlight that the results contain hash codes, which guarantee fair comparison with other results.

\noindent \textbf{Reproduction}: If a user wants to reproduce the results, the user only needs to run our tool to fetch the data uploaded by another user, including the code and the running environment.

\noindent \textbf{Dashboard}~\footnoteref{footnote:dashboard}: The dashboard is a website that maintains the results uploaded by users, which makes it convenient to compare performances of models and find state-of-the-art models. As our unique feature, users can submit the results by running the code from other users, which further facilitates reproduction. %We provide a platform to gather the contributions from the community, which is more feasible than a centralized verification system.

\begin{figure*}[!t]
  \centering
  \scalebox{1}[0.90]{
  \includegraphics[width=0.95\linewidth]{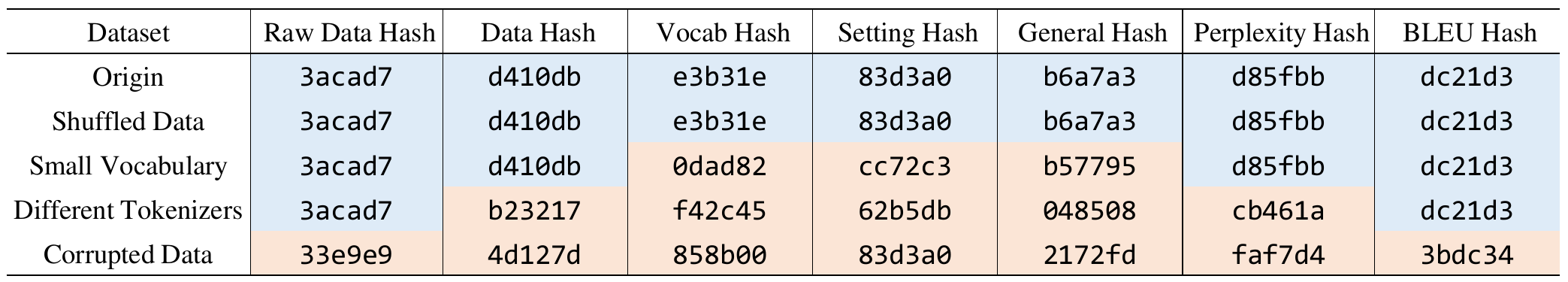}
  }
  \caption{Different hash codes (only the leading 6 characters showed) in the different settings. The table demonstrates that hash code can identify the differences in settings and avoid unfair comparisons. }%Please refer to Section \ref{sec:dataloader} and Section \ref{sec:metric} for the types of hash codes. }
  \label{fig:hashcodes}
  \vspace{-1em}
\end{figure*}

\subsection{Resources and Benchmark Models}

\label{sec:resources-baselines}

To improve usability, we further provide resources and benchmark models compatible with our toolkit. The resources include benchmark datasets, pretrained model weights and more, which can be automatically downloaded by data loaders in CoTK. Some resources and benchmark models we provide are presented in Table \ref{tab:resources} and Table \ref{tab:baselines}. %Please visit our homepage for more resources and baselines.

\subsection{Other Features}

\noindent \textbf{Batched Data:}
CoTK is specially designed for deep learning models, where all APIs receive batched data. Batched sentences can be directly converted to/from tensors. This feature avoids errors of manipulating paddings.%, which may bring errors, as shown in Figure \ref{fig:intro-example} (c). 

%\vspace{0.5em}

\noindent \textbf{Compatibility:}
CoTK is not dependent on the deep learning frameworks, such as TensorFlow \cite{tensorflow2016} and PyTorch \cite{pytorch2019}. CoTK is also compatible with many other toolkits of text generation, such as Texar \cite{texar2019} and Fairseq \cite{ott2019fairseq}, which means the model implemented with these tookits can be evaluated or published with CoTK. Moreover, the comparison across frameworks is also possible.%, where we believe differences introduced by the deep learning framework are also important. 

%\vspace{0.5em}

\noindent \textbf{Extensibility:} %
CoTK is highly extensible, where new tasks, metrics and benchmark models can be easily integrated into the toolkit. We believe that CoTK can grow with the advancements of text generation in the community.

%We are planning to collect implementations from the community, so that we can help achieve fair evaluation over more tasks.

\section{Proof-of-Concept Examples}

\label{sec:showcase}

\subsection{Hash Codes of Different Settings}

\label{sec:hashcode-example}
%As aforementioned, data loaders and metrics provides hash codes to help researchers to find whether the data is the same or the result is comparable. Here 
We present an example to demonstrate how hash codes can identify differences in settings. We choose a subset of \textit{OpenSubtitles} as our dataset (\textbf{Origin}), and modify it for four settings: 
\textbf{Shuffled data.} The lines of the data file are shuffled, which only affects the order of the samples. 
\textbf{Small vocabulary.} The size of frequent vocabulary is changed from 1323 to 752. 
\textbf{Different tokenizers.} The tokenizer is changed from the Punkt tokenizer to the tokenizer of BERT~\footnote{The tokenizers of BERT are \textit{bert-base-uncased} from \url{https://github.com/huggingface/transformers}.}.
\textbf{Corrupted data.} A sample from the dataset is removed.

The result is presented in Figure \ref{fig:hashcodes}. Shuffling data does not change any hash code because it does not affect training or evaluation. On the dataset of small vocabulary, \textit{Vocab Hash} and \textit{Setting Hash} codes are different. However, hash codes for metrics do not change since the result is still comparable. On the dataset of different tokenizers, the \textit{Perplexity Hash} is changed, because comparison under different tokenizers is not supported by perplexity. On corrupted data, all the hash codes are changed, where \textit{Raw Data Hash} signifies that it is a different dataset.

\subsection{\mbox{Comparison under Different Vocabularies}}

%As aforementioned, the original perplexity is unfair when predicting \unk under different vocabularies, and we improve the perplexity to achieve fair comparisons.
We present an example to demonstrate that we can achieve fair comparison under different vocabularies.
We train a GRU text generation model on the dataset \textit{MSCOCO} with different frequent vocabularies, and show how the vocabulary size affects the result.

\begin{table} [!htp]
\centering
\small
\setlength{\tabcolsep}{1.0mm}{
\begin{tabular}{cccc}
\hline
$T_{min}$ & $|F|$ & CoTK's Perplexity & Original Perplexity\\
\hline
1 & 30765 & 14.17 & \textcolor{deepgray}{14.15} \\
2 & 19227 & 14.11 & \textcolor{deepgray}{13.95} \\
4 & 12555 & \textbf{14.10} & \textcolor{deepgray}{13.74} \\
10 & 8044 & 14.22 & \textcolor{deepgray}{13.39} \\
%20 & 5737 & 14.91 & 13.09 \\
40 & 4062 & 15.30 & \textcolor{deepgray}{12.69} \\
%80 & 2832 & 16.08 & 12.15 \\
160 & 1900 & 17.13 & \textcolor{deepgray}{11.49} \\
\hline
\end{tabular}
}
\caption{Perplexity under different sets of the frequent vocabulary. In each row, the words appearing less than $T_{min}$ times in the training data are regarded as rare words. The results of CoTK's perplexity are comparable while those of the original perplexity are not.}
\label{tab:perplexity-vocab}
\vspace{-0.5em}
\end{table}

The result is presented in Table \ref{tab:perplexity-vocab}. When $T_{min}=4$, the model reaches the best CoTK's perplexity. However, the original perplexity will get smaller as $|F|$ decreases, and it will reach 1 when $|F|=1$. It shows that the original perplexity is not a fair metric under different vocabularies.

\subsection{Evaluation of Benchmark Models}
\label{sec:evaluation}

We demonstrate the evaluation results of some benchmark models on text generation (without input) and single-turn dialog generation, as shown in Table \ref{tab:result-language} and Table \ref{tab:result-singleturn}. The details of implementation and metrics are presented in the appendix.

Notice that the perplexity of GPT2-ft(finetune) cannot be fairly compared with the other models, because their tokenization are different. However, the other metrics, including BLEU, S-BLEU~\cite{texygen2018}, F/B/H-BLEU~\cite{irl2018}, F/R-PPL~\cite{arae2018}, Distinct-2~\cite{distinct2016li}, standardize the tokenization with the same method of BLEU described in Section \ref{sec:metric}, so the results of these metrics are comparable among the models.
%%%%%这个实验是要说，我们的结论相比其他论文的结论有所不同？

\begin{table} [!t]
\centering
\small
\setlength{\tabcolsep}{0.5mm}{
\begin{tabular}{c|ccccc}
\hline
Model & PPL & S-BLEU & F/B/H-BLEU & F/R-PPL \\
\hline
GRU & 47.74 & 30.8 & \textbf{27.5}/20.1/23.2 & 80.9/186.2 \\
Transformer& \textbf{37.04} & \textbf{30.2} & 25.3/20.3/22.5 & 80.7/180.3\\
GPT2-ft & \textcolor{deepgray}{19.30*} & 31.7 & 26.8/\textbf{20.5}/\textbf{23.3} & \textbf{76.1}/\textbf{168.9}\\
\hline
\end{tabular}
}
\caption{Evaluation results of text generation (without input) models on the \textit{EMNLP2017} dataset. (*) indicates the value is not comparable with its counterparts.}
\label{tab:result-language}
\vspace{-0.5em}
\end{table}

\begin{table} [!t]
\centering
\small
\setlength{\tabcolsep}{0.5mm}{
\begin{tabular}{c|ccccc}
\hline
Model & PPL & BLEU & Distinct-2 &  F/R-PPL \\
\hline
GRU &  \textbf{37.26} & 1.07 & 0.094 & 23.4/243.3\\
Transformer & 39.36 & 0.99 & 0.063 & \textbf{19.5}/308.6\\
GPT2-ft & \textcolor{deepgray}{21.12*} & \textbf{1.30} & \textbf{0.156} & 29.8/\textbf{124.2}\\
\hline
\end{tabular}
}
\caption{Evaluation results of single-turn dialog models on \textit{OpenSubtitles} dataset. (*) indicates the value is not comparable with other counterparts.}
\label{tab:result-singleturn}
\vspace{-0.5em}
\end{table}

\subsection{Other Examples}

%We demonstrate an example of development procedure, a comparisons 
More examples about usage, model publication and reproduction are demonstrated in the appendix.

\section{Related Work}
\label{sec:related}
%%%%注意工具包名字的大小写！
%%%%With the rapid development of deep learning in NLP, many toolkits are developed to support researchers from different aspects. 
%%%We survey related work and the unique advantages of CoTK from four aspects.
Unlike
%PyTorch \cite{pytorch2019} and Tensorflow (or Keras \footnote{\href{https://www.tensorflow.org/guide/keras}{https://www.tensorflow.org/guide/keras}}) \cite{tensorflow2016} that provides general programming frameworks for implementing deep learning models, or 
PyTorch-NLP~\cite{pytorch-nlp}, torchtext\footnote{\url{https://github.com/pytorch/text}},
AllenNLP~\cite{gardner2018allennlp}, and GluonNLP~\cite{guo2019gluoncv} that provide common modules and utilities in NLP,
%, or 
%HuggingFace's toolkit that provides Transformer-based models~\cite{Wolf2019HuggingFacesTS} for general use of NLP,
CoTK mainly focuses on text generation. 
Analogous to ours,
Texar~\cite{texar2019} and Fairseq~\cite{ott2019fairseq} provide state-of-the-art models in text generation.
However, these toolkits are largely targeting at model implementation.
%%%%%
CoTK characterizes itself by focusing more on data processing, evaluation and reproduction, but also allowing the compatibility with other toolkits: the models implemented by these toolkits can be easily evaluated with CoTK.

%%%%is not focusing on model implementation but compatible with all above toolkits. These toolkits assist the model implementation, and CoTK helps the other steps in the development procedure, including data processing, evaluation, and reproduction. The compatibility is intentionally kept to satisfy different preferences and achieve fair comparisons among different implementations. %We intentionally keep our toolkit independent on the model implementations

\vspace{0.5em}

\noindent \textbf{Data Processing} \\
Many toolkits provide data loaders as we do. Texar and Fairseq implement utilities for loading data and provide benchmarks for text generation tasks. PyTorch-NLP, torchtext and AllenNLP provide a similar function for general NLP tasks.

In comparison, as a unique feature, CoTK uses hash code to identify the differences and remind researchers when experimental settings change. Furthermore, the data loaders work with our implemented metrics to realize fair comparison across different settings and datasets.

\vspace{0.5em}

\noindent \textbf{Evaluation} \\
The evaluation of text generation are less touched by previous toolkits. 
Most of toolkits, such as torchtext, PyTorch-NLP and Texar, only provide few metrics like BLEU. Although NLTK \cite{loper2002nltk} and AllenNLP contain evaluation modules, few of which are designed for text generation. 

We provide unified APIs to receive batched samples, which are convenient for deep learning models. Moreover, hash code plays an essential role in our toolkit to achieve fair comparison.

\vspace{0.5em}

\noindent \textbf{Publication and Reproduction} \\
Publication and reproduction are rarely addressed by existing toolkits for text generation. Here we list two applications to achieve a similar function.
%: sacred\cite{greff2017sacred} and a website named \textit{Paper with Code}\footnote{\href{https://paperswithcode.com/}{https://paperswithcode.com/}}. 
Sacred \cite{greff2017sacred} is an experiment management tool, where the configurations, codes and results are tracked for reproduction. It is only for individuals and not designed for sharing results with the community. \textit{Paper with Code}\footnote{\url{https://paperswithcode.com/}} collects the evaluation results and leaderboards for different tasks. However, the results are manually filled and can be hard to reproduce. CoTK can track the codes, the results and the running environment automatically and publish them to the community, which is more convenient and efficient for comparison and reproducibility.

\section{Conclusion and Future Work}

In this paper, we introduce CoTK, a toolkit for fast development and fair evaluation of text generation. 
CoTK provides support through the entire lifetime of model development and addresses issues that are often ignored but lead to unfair comparison. With CoTK, researchers can easily handle data processing, model evaluation, and reproduction. Our special design signifies when and which metric cannot be fairly compared.
%It is also possible to use other toolkits to implement models, where CoTK is compatible with the toolkits to provide evaluation or other functionalities.
CoTK can grow with the development of text generation in the community where more tasks, metrics, resources and benchmark models can be constantly integrated into our toolkit. 
%%%%We also invite researchers to join our toolkit, evaluate their models fairly, and publish the models to the community. We believe these actions can push forward the text generation research in the end.
We believe that this toolkit will not only facilitate researchers to develop text generation models, but also support fair comparison among models and promote the reproducibility of these models.

\bibliography{acl2020}

\begin{thebibliography}{42}
\expandafter\ifx\csname natexlab\endcsname\relax\def\natexlab#1{#1}\fi

\bibitem[{Abadi et~al.(2016)Abadi, Barham, Chen, Chen, Davis, Dean, Devin,
  Ghemawat, Irving, Isard, Kudlur, Levenberg, Monga, Moore, Murray, Steiner,
  Tucker, Vasudevan, Warden, Wicke, Yu, and Zheng}]{tensorflow2016}
Mart{\'{\i}}n Abadi, Paul Barham, Jianmin Chen, Zhifeng Chen, Andy Davis,
  Jeffrey Dean, Matthieu Devin, Sanjay Ghemawat, Geoffrey Irving, Michael
  Isard, Manjunath Kudlur, Josh Levenberg, Rajat Monga, Sherry Moore,
  Derek~Gordon Murray, Benoit Steiner, Paul~A. Tucker, Vijay Vasudevan, Pete
  Warden, Martin Wicke, Yuan Yu, and Xiaoqiang Zheng. 2016.
\newblock \href
  {https://www.usenix.org/conference/osdi16/technical-sessions/presentation/abadi}
  {Tensorflow: {A} system for large-scale machine learning}.
\newblock In \emph{12th {USENIX} Symposium on Operating Systems Design and
  Implementation, {OSDI} 2016, Savannah, GA, USA, November 2-4, 2016}, pages
  265--283.

\bibitem[{Ahn et~al.(2016)Ahn, Choi, P{\"a}rnamaa, and Bengio}]{ahn2016upp}
Sungjin Ahn, Heeyoul Choi, Tanel P{\"a}rnamaa, and Yoshua Bengio. 2016.
\newblock A neural knowledge language model.
\newblock \emph{arXiv preprint arXiv:1608.00318}.

\bibitem[{Banerjee and Lavie(2005)}]{meteor2005banerjee}
Satanjeev Banerjee and Alon Lavie. 2005.
\newblock \href {https://www.aclweb.org/anthology/W05-0909/} {{METEOR:} an
  automatic metric for {MT} evaluation with improved correlation with human
  judgments}.
\newblock In \emph{Proceedings of the Workshop on Intrinsic and Extrinsic
  Evaluation Measures for Machine Translation and/or Summarization@ACL 2005,
  Ann Arbor, Michigan, USA, June 29, 2005}, pages 65--72.

\bibitem[{Brown et~al.(1992)Brown, Pietra, Pietra, Lai, and
  Mercer}]{perplexity1992brown}
Peter~F. Brown, Stephen~Della Pietra, Vincent J.~Della Pietra, Jennifer~C. Lai,
  and Robert~L. Mercer. 1992.
\newblock An estimate of an upper bound for the entropy of english.
\newblock \emph{Computational Linguistics}, 18(1):31--40.

\bibitem[{Chen et~al.(2015)Chen, Fang, Lin, Vedantam, Gupta, Doll{\'a}r, and
  Zitnick}]{chen2015mscoco}
Xinlei Chen, Hao Fang, Tsung-Yi Lin, Ramakrishna Vedantam, Saurabh Gupta, Piotr
  Doll{\'a}r, and C~Lawrence Zitnick. 2015.
\newblock Microsoft coco captions: Data collection and evaluation server.
\newblock \emph{arXiv preprint arXiv:1504.00325}.

\bibitem[{Chung et~al.(2014)Chung, Gulcehre, Cho, and Bengio}]{gru2014chung}
Junyoung Chung, Caglar Gulcehre, KyungHyun Cho, and Yoshua Bengio. 2014.
\newblock Empirical evaluation of gated recurrent neural networks on sequence
  modeling.
\newblock \emph{arXiv preprint arXiv:1412.3555}.

\bibitem[{Devlin et~al.(2019)Devlin, Chang, Lee, and Toutanova}]{bert2019}
Jacob Devlin, Ming{-}Wei Chang, Kenton Lee, and Kristina Toutanova. 2019.
\newblock \href {https://www.aclweb.org/anthology/N19-1423/} {{BERT:}
  pre-training of deep bidirectional transformers for language understanding}.
\newblock In \emph{Proceedings of the 2019 Conference of the North American
  Chapter of the Association for Computational Linguistics: Human Language
  Technologies, {NAACL-HLT} 2019, Minneapolis, MN, USA, June 2-7, 2019, Volume
  1 (Long and Short Papers)}, pages 4171--4186.

\bibitem[{Forgues et~al.(2014)Forgues, Pineau, Larchev{\^e}que, and
  Tremblay}]{forgues2014bootstrapping}
Gabriel Forgues, Joelle Pineau, Jean-Marie Larchev{\^e}que, and R{\'e}al
  Tremblay. 2014.
\newblock Bootstrapping dialog systems with word embeddings.
\newblock In \emph{Nips, modern machine learning and natural language
  processing workshop}, volume~2.

\bibitem[{Gardner et~al.(2018)Gardner, Grus, Neumann, Tafjord, Dasigi, Liu,
  Peters, Schmitz, and Zettlemoyer}]{gardner2018allennlp}
Matt Gardner, Joel Grus, Mark Neumann, Oyvind Tafjord, Pradeep Dasigi, Nelson
  Liu, Matthew Peters, Michael Schmitz, and Luke Zettlemoyer. 2018.
\newblock Allennlp: A deep semantic natural language processing platform.
\newblock \emph{arXiv preprint arXiv:1803.07640}.

\bibitem[{Graves(2013)}]{graves2013rnn}
Alex Graves. 2013.
\newblock Generating sequences with recurrent neural networks.
\newblock \emph{arXiv preprint arXiv: 1308.0850}.

\bibitem[{Greff et~al.(2017)Greff, Klein, Chovanec, Hutter, and
  Schmidhuber}]{greff2017sacred}
Klaus Greff, Aaron Klein, Martin Chovanec, Frank Hutter, and J{\"u}rgen
  Schmidhuber. 2017.
\newblock The sacred infrastructure for computational research.
\newblock In \emph{Proceedings of the Python in Science Conferences-SciPy
  Conferences}.

\bibitem[{Guo et~al.(2019)Guo, He, He, Lausen, Li, Lin, Shi, Wang, Xie, Zha
  et~al.}]{guo2019gluoncv}
Jian Guo, He~He, Tong He, Leonard Lausen, Mu~Li, Haibin Lin, Xingjian Shi,
  Chenguang Wang, Junyuan Xie, Sheng Zha, et~al. 2019.
\newblock Gluoncv and gluonnlp: Deep learning in computer vision and natural
  language processing.
\newblock \emph{arXiv preprint arXiv:1907.04433}.

\bibitem[{He et~al.(2017)He, Liu, Liu, and Zhao}]{copynet2017he}
Shizhu He, Cao Liu, Kang Liu, and Jun Zhao. 2017.
\newblock \href {https://doi.org/10.18653/v1/P17-1019} {Generating natural
  answers by incorporating copying and retrieving mechanisms in
  sequence-to-sequence learning}.
\newblock In \emph{Proceedings of the 55th Annual Meeting of the Association
  for Computational Linguistics, {ACL} 2017, Vancouver, Canada, July 30 -
  August 4, Volume 1: Long Papers}, pages 199--208.

\bibitem[{Hu et~al.(2019)Hu, Shi, Tan, Wang, Yang, Zhao, He, Qin, Wang, Ma,
  Liu, Liang, Zhu, Sachan, and Xing}]{texar2019}
Zhiting Hu, Haoran Shi, Bowen Tan, Wentao Wang, Zichao Yang, Tiancheng Zhao,
  Junxian He, Lianhui Qin, Di~Wang, Xuezhe Ma, Zhengzhong Liu, Xiaodan Liang,
  Wanrong Zhu, Devendra~Singh Sachan, and Eric~P. Xing. 2019.
\newblock \href {https://www.aclweb.org/anthology/P19-3027/} {Texar: {A}
  modularized, versatile, and extensible toolkit for text generation}.
\newblock In \emph{Proceedings of the 57th Conference of the Association for
  Computational Linguistics, {ACL} 2019, Florence, Italy, July 28 - August 2,
  2019, Volume 3: System Demonstrations}, pages 159--164.

\bibitem[{Huang et~al.(2019)Huang, Zhu, and Gao}]{huang2019challenges}
Minlie Huang, Xiaoyan Zhu, and Jianfeng Gao. 2019.
\newblock Challenges in building intelligent open-domain dialog systems.
\newblock \emph{arXiv preprint arXiv:1905.05709}.

\bibitem[{Kingma and Welling(2014)}]{vae2014}
Diederik~P. Kingma and Max Welling. 2014.
\newblock \href {http://arxiv.org/abs/1312.6114} {Auto-encoding variational
  bayes}.
\newblock In \emph{2nd International Conference on Learning Representations,
  {ICLR} 2014, Banff, AB, Canada, April 14-16, 2014, Conference Track
  Proceedings}.

\bibitem[{Kiss and Strunk(2006)}]{Punkt}
Tibor Kiss and Jan Strunk. 2006.
\newblock \href {https://doi.org/10.1162/coli.2006.32.4.485} {Unsupervised
  multilingual sentence boundary detection}.
\newblock \emph{Computational Linguistics}, 32(4):485--525.

\bibitem[{Li et~al.(2016)Li, Galley, Brockett, Gao, and Dolan}]{distinct2016li}
Jiwei Li, Michel Galley, Chris Brockett, Jianfeng Gao, and Bill Dolan. 2016.
\newblock \href {https://www.aclweb.org/anthology/N16-1014/} {A
  diversity-promoting objective function for neural conversation models}.
\newblock In \emph{{NAACL} {HLT} 2016, The 2016 Conference of the North
  American Chapter of the Association for Computational Linguistics: Human
  Language Technologies, San Diego California, USA, June 12-17, 2016}, pages
  110--119.

\bibitem[{Lin(2004)}]{rouge2014lin}
Chin-Yew Lin. 2004.
\newblock \href {https://www.aclweb.org/anthology/W04-1013} {{ROUGE}: A package
  for automatic evaluation of summaries}.
\newblock In \emph{Text Summarization Branches Out}, pages 74--81, Barcelona,
  Spain. Association for Computational Linguistics.

\bibitem[{Loper and Bird(2002)}]{loper2002nltk}
Edward Loper and Steven Bird. 2002.
\newblock Nltk: the natural language toolkit.
\newblock \emph{arXiv preprint cs/0205028}.

\bibitem[{Lowe et~al.(2015)Lowe, Pow, Serban, and Pineau}]{ubuntu2015lowe}
Ryan Lowe, Nissan Pow, Iulian Serban, and Joelle Pineau. 2015.
\newblock \href {https://www.aclweb.org/anthology/W15-4640/} {The ubuntu
  dialogue corpus: {A} large dataset for research in unstructured multi-turn
  dialogue systems}.
\newblock In \emph{Proceedings of the {SIGDIAL} 2015 Conference, The 16th
  Annual Meeting of the Special Interest Group on Discourse and Dialogue, 2-4
  September 2015, Prague, Czech Republic}, pages 285--294.

\bibitem[{Ott et~al.(2019)Ott, Edunov, Baevski, Fan, Gross, Ng, Grangier, and
  Auli}]{ott2019fairseq}
Myle Ott, Sergey Edunov, Alexei Baevski, Angela Fan, Sam Gross, Nathan Ng,
  David Grangier, and Michael Auli. 2019.
\newblock fairseq: A fast, extensible toolkit for sequence modeling.
\newblock In \emph{Proceedings of NAACL-HLT 2019: Demonstrations}.

\bibitem[{Papineni et~al.(2002)Papineni, Roukos, Ward, and
  Zhu}]{bleu2002papineni}
Kishore Papineni, Salim Roukos, Todd Ward, and Wei{-}Jing Zhu. 2002.
\newblock \href {https://www.aclweb.org/anthology/P02-1040/} {Bleu: a method
  for automatic evaluation of machine translation}.
\newblock In \emph{Proceedings of the 40th Annual Meeting of the Association
  for Computational Linguistics, July 6-12, 2002, Philadelphia, PA, {USA}},
  pages 311--318.

\bibitem[{Paszke et~al.(2019)Paszke, Gross, Massa, Lerer, Bradbury, Chanan,
  Killeen, Lin, Gimelshein, Antiga, Desmaison, K{\"{o}}pf, Yang, DeVito,
  Raison, Tejani, Chilamkurthy, Steiner, Fang, Bai, and Chintala}]{pytorch2019}
Adam Paszke, Sam Gross, Francisco Massa, Adam Lerer, James Bradbury, Gregory
  Chanan, Trevor Killeen, Zeming Lin, Natalia Gimelshein, Luca Antiga, Alban
  Desmaison, Andreas K{\"{o}}pf, Edward Yang, Zachary DeVito, Martin Raison,
  Alykhan Tejani, Sasank Chilamkurthy, Benoit Steiner, Lu~Fang, Junjie Bai, and
  Soumith Chintala. 2019.
\newblock \href
  {http://papers.nips.cc/paper/9015-pytorch-an-imperative-style-high-performance-deep-learning-library}
  {Pytorch: An imperative style, high-performance deep learning library}.
\newblock In \emph{Advances in Neural Information Processing Systems 32: Annual
  Conference on Neural Information Processing Systems 2019, NeurIPS 2019, 8-14
  December 2019, Vancouver, BC, Canada}, pages 8024--8035.

\bibitem[{Petrochuk(2018)}]{pytorch-nlp}
Michael Petrochuk. 2018.
\newblock Pytorch-nlp: Rapid prototyping with pytorch natural language
  processing (nlp) tools.
\newblock \url{https://github.com/PetrochukM/PyTorch-NLP}.

\bibitem[{Radford et~al.(2019)Radford, Wu, Child, Luan, Amodei, and
  Sutskever}]{gpt22019radford}
Alec Radford, Jeffrey Wu, Rewon Child, David Luan, Dario Amodei, and Ilya
  Sutskever. 2019.
\newblock Language models are unsupervised multitask learners.
\newblock \emph{OpenAI Blog}, 1(8).

\bibitem[{Roemmele(2016)}]{story2016roemmele}
Melissa Roemmele. 2016.
\newblock \href
  {http://www.aaai.org/ocs/index.php/AAAI/AAAI16/paper/view/11966} {Writing
  stories with help from recurrent neural networks}.
\newblock In \emph{Proceedings of the Thirtieth {AAAI} Conference on Artificial
  Intelligence, February 12-17, 2016, Phoenix, Arizona, {USA}}, pages
  4311--4342.

\bibitem[{Rush et~al.(2015)Rush, Chopra, and Weston}]{summarization2015rush}
Alexander~M. Rush, Sumit Chopra, and Jason Weston. 2015.
\newblock \href {https://www.aclweb.org/anthology/D15-1044/} {A neural
  attention model for abstractive sentence summarization}.
\newblock In \emph{Proceedings of the 2015 Conference on Empirical Methods in
  Natural Language Processing, {EMNLP} 2015, Lisbon, Portugal, September 17-21,
  2015}, pages 379--389.

\bibitem[{Shi et~al.(2018)Shi, Chen, Qiu, and Huang}]{irl2018}
Zhan Shi, Xinchi Chen, Xipeng Qiu, and Xuanjing Huang. 2018.
\newblock \href {https://doi.org/10.24963/ijcai.2018/606} {Toward diverse text
  generation with inverse reinforcement learning}.
\newblock In \emph{Proceedings of the Twenty-Seventh International Joint
  Conference on Artificial Intelligence, {IJCAI} 2018, July 13-19, 2018,
  Stockholm, Sweden.}, pages 4361--4367.

\bibitem[{Sordoni et~al.(2015)Sordoni, Bengio, Vahabi, Lioma, Simonsen, and
  Nie}]{hred2015sordoni}
Alessandro Sordoni, Yoshua Bengio, Hossein Vahabi, Christina Lioma, Jakob~Grue
  Simonsen, and Jian{-}Yun Nie. 2015.
\newblock \href {https://doi.org/10.1145/2806416.2806493} {A hierarchical
  recurrent encoder-decoder for generative context-aware query suggestion}.
\newblock In \emph{Proceedings of the 24th {ACM} International Conference on
  Information and Knowledge Management, {CIKM} 2015, Melbourne, VIC, Australia,
  October 19 - 23, 2015}, pages 553--562.

\bibitem[{Sutskever et~al.(2011)Sutskever, Martens, and
  Hinton}]{text2011sutskever}
Ilya Sutskever, James Martens, and Geoffrey~E. Hinton. 2011.
\newblock \href {https://icml.cc/2011/papers/524\_icmlpaper.pdf} {Generating
  text with recurrent neural networks}.
\newblock In \emph{Proceedings of the 28th International Conference on Machine
  Learning, {ICML} 2011, Bellevue, Washington, USA, June 28 - July 2, 2011},
  pages 1017--1024.

\bibitem[{Sutskever et~al.(2014)Sutskever, Vinyals, and
  Le}]{seq2seq2014sutskever}
Ilya Sutskever, Oriol Vinyals, and Quoc~V. Le. 2014.
\newblock \href
  {http://papers.nips.cc/paper/5346-sequence-to-sequence-learning-with-neural-networks}
  {Sequence to sequence learning with neural networks}.
\newblock In \emph{Advances in Neural Information Processing Systems 27: Annual
  Conference on Neural Information Processing Systems 2014, December 8-13 2014,
  Montreal, Quebec, Canada}, pages 3104--3112.

\bibitem[{Tiedemann(2016)}]{opensubtitles2016tiedemann}
J{\"o}rg Tiedemann. 2016.
\newblock Finding alternative translations in a large corpus of movie subtitle.
\newblock In \emph{Proceedings of the Tenth International Conference on
  Language Resources and Evaluation (LREC'16)}, pages 3518--3522.

\bibitem[{Vaswani et~al.(2017)Vaswani, Shazeer, Parmar, Uszkoreit, Jones,
  Gomez, Kaiser, and Polosukhin}]{vaswano17transformer}
Ashish Vaswani, Noam Shazeer, Niki Parmar, Jakob Uszkoreit, Llion Jones,
  Aidan~N. Gomez, Lukasz Kaiser, and Illia Polosukhin. 2017.
\newblock Attention is all you need.
\newblock In \emph{Advances in Neural Information Processing Systems}, pages
  5998--6008.

\bibitem[{Vinyals and Le(2015)}]{vinyals2015conversational}
Oriol Vinyals and Quoc Le. 2015.
\newblock A neural conversational model.
\newblock \emph{arXiv preprint arXiv:1506.05869}.

\bibitem[{Vinyals et~al.(2015)Vinyals, Toshev, Bengio, and
  Erhan}]{imagecaption2015vinyals}
Oriol Vinyals, Alexander Toshev, Samy Bengio, and Dumitru Erhan. 2015.
\newblock \href {https://doi.org/10.1109/CVPR.2015.7298935} {Show and tell: {A}
  neural image caption generator}.
\newblock In \emph{{IEEE} Conference on Computer Vision and Pattern
  Recognition, {CVPR} 2015, Boston, MA, USA, June 7-12, 2015}, pages
  3156--3164.

\bibitem[{Wolf et~al.(2019)Wolf, Sanh, Chaumond, and
  Delangue}]{wolf2019transfertransfo}
Thomas Wolf, Victor Sanh, Julien Chaumond, and Clement Delangue. 2019.
\newblock Transfertransfo: A transfer learning approach for neural network
  based conversational agents.
\newblock \emph{arXiv preprint arXiv:1901.08149}.

\bibitem[{Yang et~al.(2019)Yang, Dai, Yang, Carbonell, Salakhutdinov, and
  Le}]{xlnet2019yang}
Zhilin Yang, Zihang Dai, Yiming Yang, Jaime~G. Carbonell, Ruslan Salakhutdinov,
  and Quoc~V. Le. 2019.
\newblock \href
  {http://papers.nips.cc/paper/8812-xlnet-generalized-autoregressive-pretraining-for-language-understanding}
  {Xlnet: Generalized autoregressive pretraining for language understanding}.
\newblock In \emph{Advances in Neural Information Processing Systems 32: Annual
  Conference on Neural Information Processing Systems 2019, NeurIPS 2019, 8-14
  December 2019, Vancouver, BC, Canada}, pages 5754--5764.

\bibitem[{Zhao et~al.(2018)Zhao, Kim, Zhang, Rush, and LeCun}]{arae2018}
Junbo~Jake Zhao, Yoon Kim, Kelly Zhang, Alexander~M. Rush, and Yann LeCun.
  2018.
\newblock \href {http://proceedings.mlr.press/v80/zhao18b.html} {Adversarially
  regularized autoencoders}.
\newblock In \emph{Proceedings of the 35th International Conference on Machine
  Learning, {ICML} 2018, Stockholmsm{\"{a}}ssan, Stockholm, Sweden, July 10-15,
  2018}, pages 5897--5906.

\bibitem[{Zhao et~al.(2017)Zhao, Zhao, and Esk{\'{e}}nazi}]{cvae2017}
Tiancheng Zhao, Ran Zhao, and Maxine Esk{\'{e}}nazi. 2017.
\newblock \href {https://doi.org/10.18653/v1/P17-1061} {Learning
  discourse-level diversity for neural dialog models using conditional
  variational autoencoders}.
\newblock In \emph{Proceedings of the 55th Annual Meeting of the Association
  for Computational Linguistics, {ACL} 2017, Vancouver, Canada, July 30 -
  August 4, Volume 1: Long Papers}, pages 654--664.

\bibitem[{Zhou et~al.(2018)Zhou, Huang, Zhang, Zhu, and Liu}]{ecm2018zhou}
Hao Zhou, Minlie Huang, Tianyang Zhang, Xiaoyan Zhu, and Bing Liu. 2018.
\newblock \href
  {https://www.aaai.org/ocs/index.php/AAAI/AAAI18/paper/view/16455} {Emotional
  chatting machine: Emotional conversation generation with internal and
  external memory}.
\newblock In \emph{Proceedings of the Thirty-Second {AAAI} Conference on
  Artificial Intelligence, (AAAI-18), the 30th innovative Applications of
  Artificial Intelligence (IAAI-18), and the 8th {AAAI} Symposium on
  Educational Advances in Artificial Intelligence (EAAI-18), New Orleans,
  Louisiana, USA, February 2-7, 2018}, pages 730--739.

\bibitem[{Zhu et~al.(2018)Zhu, Lu, Zheng, Guo, Zhang, Wang, and
  Yu}]{texygen2018}
Yaoming Zhu, Sidi Lu, Lei Zheng, Jiaxian Guo, Weinan Zhang, Jun Wang, and Yong
  Yu. 2018.
\newblock \href {https://doi.org/10.1145/3209978.3210080} {Texygen: {A}
  benchmarking platform for text generation models}.
\newblock In \emph{The 41st International {ACM} {SIGIR} Conference on Research
  {\&} Development in Information Retrieval, {SIGIR} 2018, Ann Arbor, MI, USA,
  July 08-12, 2018}, pages 1097--1100.

\end{thebibliography}
\bibliographystyle{acl_natbib}

%\begin{comment}
\appendix

\section{An Example of Development Procedure}

\begin{figure*}[!t]
  \centering
  %\scalebox{1}[0.95]{
  \includegraphics[width=\linewidth]{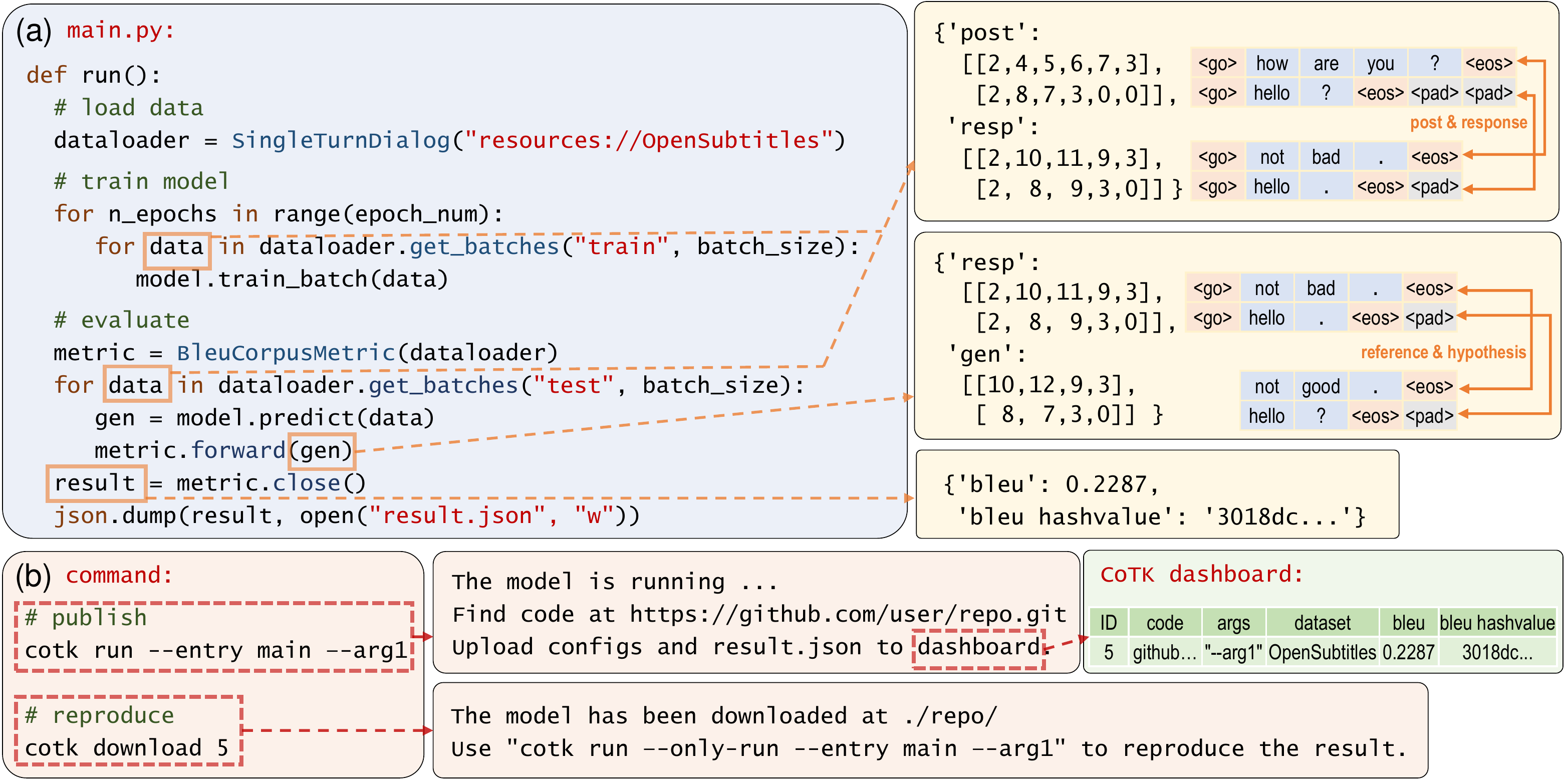}
  %}
  \caption{An example of a development procedure with the help of CoTK. (a) The file "main.py", including codes for loading data, training model, and performing evaluation. The implementation of the model are omitted. (b) Two commands for publishing and reproducing results. }
  \label{fig:development-example}
  \vspace{-1em}
\end{figure*} 

\label{sec:example-procedure}

With the help of CoTK, it is convenient to develop a novel model, as an example of the development procedure shown in Figure \ref{fig:development-example}. 

In (a), we show the main code for data processing, model training and evaluation. The three parts of code are explained as follows:
\begin{itemize}[leftmargin=1em]
    \setlength{\itemsep}{0ex}
    \setlength{\parskip}{2px}
    \vspace{-0.4em}

\item \textbf{Data processing}. The dataset is downloaded and processed in a single line, where \textit{SingleTurnDialog} is a data loader (Section \ref{sec:dataloader}) %\ref{sec:dataloader}
and \textit{OpenSubtitles} is a resource (Section \ref{sec:resources-baselines}). % \ref{sec:resources-baselines})

\item \textbf{Model training}. Our data loader provides batches of data, where sentences are converted to index and padded. This format is commonly used by most of the text generation models.
%The batched data is convenient to be converted into tensor in Tensorflow or PyTorch. 
%
\item \textbf{Model evaluation}. The model can generate data in batches, which are passed to a metric (Section ref{sec:metric})
, \textit{BleuCorpusMetric}, in our case. The metric object produces the result with hash codes.
 %\vspace{-1em}
\end{itemize}

In (b), we show two commands for publishing and reproducing results, respectively. The first command uploads the code to Github, the running environment and the results to our dashboard. The second command shows how to access and reproduce the results in one line. % Please refer to Section \ref{sec:pub-repro} for more details.

\section{An Example of Comparison Under Different Tokenizers}
\label{sec:diff-tokenize}

As aforementioned, our implementation of BLEU supports fair comparison on the dataset with different tokenizers.
Here we train a GRU seq2seq model on the dataset \textit{OpenSubtitles}  with Punkt tokenizer \cite{Punkt}, and the tokenizers\footnotemark of BERT~\cite{bert2019}, GPT2~\cite{gpt22019radford} and XLNet~\cite{xlnet2019yang}. The result is presented in Table \ref{tab:diff-tokenize}. These scores are directly comparable with our implementation.

\footnotetext{\label{footnote:tokenizers}The tokenizers of BERT, GPT2, XLNet are \textit{bert-base-uncased}, \textit{gpt2}, \textit{xlnet-base-cased} respectively, from \url{https://github.com/huggingface/transformers}.}

\begin{table} [!h]
\centering
\small
\setlength{\tabcolsep}{1.0mm}{
\begin{tabular}{c|ccccc}
\hline
Tokenizer & Punkt & BERT & GPT2 & XLNet\\
\hline
BLEU-4 & \textbf{1.07} & \textbf{1.07} & 1.05 & 0.99 \\
\hline
\end{tabular} %%%%
}
\caption{BLEU-4 with different tokenizers. These values have the same hash code (3f2b67...), which indicates that the results are comparable. }
\label{tab:diff-tokenize}
\vspace{-1em}
\end{table}

%As shown in Table \ref{tab:diff-tokenize}, the tokenizer of BERT achieves best performance on BLEU-4, while Puckt tokenizer performs worst.

\section{Evaluation Details of Benchmark Models}

\subsection{Metrics}

Here we present the details of the metrics in the evaluation of text generation (without input) and single-turn dialog tasks. All the implementation can be found in the code of CoTK.

\begin{itemize}[leftmargin=1em]
    \setlength{\itemsep}{0ex}
    \setlength{\parskip}{2px}
    
    \item PPL(Perplexity)~\cite{perplexity1992brown}. Perplexity is a common metric for text generation models.
    \item BLEU~\cite{bleu2002papineni}. The metric used in the single-turn dialog task shows the overlap between generated responses and ground-truth responses. We use BLEU-4.
    \item S-BLEU(Self-BLEU)~\cite{texygen2018}. The metric used in the text generation (without input) task shows the diversity of generated sentences. We adopt BLEU-4 and use 1,000 sentences as samples.
    \item F/B/H-BLEU(Forward/Backward/Harmony BLEU)~\cite{irl2018}. The metric used in the text generation (without input) task shows fluency, diversity and overall quality of generated sentences, respectively. We adopt BLEU-4 and use 1,000 sentences as references.
    \item F/R-PPL(Forward/Reverse Perplexity)~\cite{arae2018}. The metric used in both tasks shows the fluency/diversity of generated sentences. We adopt a 5-gram language model with Kneser–Ney smoothing trained on the test set and use 10,000 sentences as references.
    \item Distinct N-gram \cite{distinct2016li}. The metric used in the single-turn dialog task shows the diversity of generated sentences. We use Distinct-2.
\end{itemize}

\subsection{Benchmark Models}

The implementation details of benchmark models for text generation (without input) and single-turn dialog tasks are presented in Table \ref{tab:details-text} and Table \ref{tab:details-dialog}, respectively. All the implementations are publicly available\footnote{\url{https://thu-coai.github.io/cotk_docs/index.html\#model-zoo}}.

\begin{table} [!htp]
\centering
\small
\setlength{\tabcolsep}{1mm}{
\begin{tabular}{c|l|lccc}
\hline
\textbf{Model} & \multicolumn{2}{c}{\textbf{Parameters}} \\
\hline
\multirow{5}{*}{GRU} & Embedding Size & 300 \\
& Decoder Features & 300 \\
& Optimizer/Learning Rate & Adam/1e-3 \\
& Decoding Strategy & Random Sampling \\
& Decoding Temperature & 0.9 \\
\hline
\multirow{7}{*}{Transformer} & Embedding Size & 300 \\
& Decoder Features & 256 \\
& Decoder Heads/Layers & 4/5 \\
& Optimizer/Learning Rate & RAdam/1e-3 \\
& Decoding Strategy & Random Sampling \\
& Decoding Temperature & 0.9 \\
\hline
\multirow{4}{*}{GPT-ft} & Pretrained Model & gpt2-117M \\
& Optimizer/Learning rate & RAdam/1e-4 \\
& Decoding Strategy & Random Sampling \\
& Decoding Temperature & 0.9 \\
\hline
\end{tabular}
}
\caption{Implementation details of text generation models (without input). }
\label{tab:details-text}
\vspace{-0.5em}
\end{table}

\begin{table} [!htp]
\centering
\small
\setlength{\tabcolsep}{1mm}{
\begin{tabular}{c|l|lccc}
\hline
\textbf{Model} & \multicolumn{2}{c}{\textbf{Parameters}} \\
\hline
\multirow{7}{*}{GRU} & Embedding Size & 300 \\
& Encoder Features & 200 (bidirectional) \\
& Decoder Features & 300 \\
& Optimizer/learning Rate & Adam/1e-3 \\
& Decoding Strategy & Top-10 Sampling \\
& Decoding Temperature & 0.9 \\
\hline
\multirow{9}{*}{Transformer} & Embedding Size & 300 \\
& Encoder Features & 256 \\
& Encoder Heads/Layers & 4/5 \\
& Decoder Features & 256 \\
& Decoder Heads/Layers & 4/5 \\
& Optimizer/learning Rate & RAdam/1e-3 \\
& Decoding Strategy & Top-10 Sampling \\
& Decoding Temperature & 0.9 \\
\hline
\multirow{4}{*}{GPT-ft} & Pretrained Model & gpt2-117M \\
& Optimizer/learning rate & RAdam/1e-4 \\
& Decoding Strategy & Top-10 Sampling \\
& Decoding Temperature & 0.9 \\
\hline
\end{tabular}
}
\caption{Implementation details of single-turn dialog generation models. }
\label{tab:details-dialog}
\vspace{-0.5em}
\end{table}

%\end{comment}

\end{document}